\documentclass[11pt]{article}
\usepackage{coling2020}
\usepackage{times}
\usepackage{url}
\usepackage{latexsym}
\usepackage{colortbl}
\usepackage{soul}
\soulregister\cite7
\soulregister\shortcite7 
\soulregister\ref7 
\usepackage{graphicx}
\definecolor{dGray}{gray}{.6}
\definecolor{mGray}{gray}{.9}
\definecolor{lGray}{gray}{.94}

\colingfinalcopy 

\title{Complaint Identification in Social Media with Transformer Networks}

\author{Mali Jin \qquad Nikolaos Aletras \\
  Department of Computer Science, University of Sheffield \\
  {\tt \{mjin6, n.aletras\}@sheffield.ac.uk} \\} 

\date{}

\begin{document}
\maketitle
\begin{abstract}

  Complaining is a speech act extensively used by humans to communicate a negative inconsistency between reality and expectations. Previous work on automatically identifying complaints in social media has focused on using feature-based and task-specific neural network models. Adapting state-of-the-art pre-trained neural language models and their combinations with other linguistic information from topics or sentiment for complaint prediction has yet to be explored. In this paper, we evaluate a battery of neural models underpinned by transformer networks which we subsequently combine with linguistic information. Experiments on a publicly available data set of complaints demonstrate that our models outperform previous state-of-the-art methods by a large margin achieving a macro F1 up to 87.
  
\end{abstract}

\section{Introduction}
\label{intro}


\blfootnote{
    
    
    
    
    \hspace{-0.65cm}  
    This work is licensed under a Creative Commons 
    Attribution 4.0 International Licence.
    Licence details:
    \url{http://creativecommons.org/licenses/by/4.0/}.
    
    
}

Complaining is a basic speech act, usually triggered by a discrepancy between reality and expectations towards an entity or event~\cite{olshtain1985complaints,cohen1993production,kowalski1996complaints}. Social media has become a popular platform for expressing complaints online \cite{preotiuc2019automatically} where customers can directly address companies regarding issues with services and products. Complaint detection aims to identify a breach of expectations in a given text snippet. However, the use of implicit and ironic expressions and accompaniment of other speech acts such as suggestions, criticism, warnings and threats~\cite{pawar2015deciphering} make it a challenging task. Identifying and classifying complaints automatically is important for: (a) improving customer service chatbots~\cite{coussement2008improving,lailiyah2017sentiment,yang2019detecting};  (b) linguists to analyze complaint characteristics on large scale \cite{vasquez2011complaints,kakolaki2016gender}; and (c) psychologists to understand the behavior of humans that express complaints~\cite{sparks2010complaining}.

Previous work has focused on binary classification between complaints and non-complaints in various domains~\cite{preotiuc2019automatically,jin2013service,coussement2008improving}. Furthermore, some studies have performed more fine-grained complaint classification. For instance, complaints directed to public authorities have been categorized based on their topics~\cite{forster2017cognitive,merson2017text} or the responsible departments~\cite{laksana2014indonesian,gunawan2018building,tjandra2015determining}. Other categorizations are based on possible hazards and risks~\cite{bhat2017identifying} as well as escalation likelihood~\cite{yang2019detecting}. Most of these previous studies have used supervised machine learning models with features extracted from text (e.g. bag-of-words, topics, features extracted from psycho-linguistic dictionaries) or task-specific neural models trained from scratch. Adapting state-of-the-art pre-trained neural language models based on transformer networks~\cite{Vaswani2017} such as BERT~\cite{devlin2018bert} and XLNet~\cite{yang2019xlnet} has yet to be explored.  

In this paper, we focus on the binary classification of Twitter posts into complaints or not \shortcite{preotiuc2019automatically}. We adapt and evaluate a battery of pre-trained transformers which we subsequently combine with external linguistic information from topics and emotions. 

\paragraph{Contributions} (1) New state-of-the-art results on complaint identification in Twitter, improving macro FI by 8.0\% over previous work by  Preotiuc-Pietro et al. \shortcite{preotiuc2019automatically}; (2) A qualitative analysis of the limitations of transformers in predicting accurately whether a given text is a complaint or not.

\section{Complaint Prediction Task and Data}

Given a text snippet (i.e. tweet), we aim to classify it as a complaint or not. For that purpose, we use the data set\footnote{\url{https://github.com/danielpreotiuc/complaints-social-media}} by Preotiuc-Pietro et al. \shortcite{preotiuc2019automatically} which contains tweets written in English that were manually annotated as complaints or not. It includes 1,232 complaints (62.4\%) and 739 non-complaints (37.6\%) over 9 domains (e.g. food, technology, etc.). Data statistics are shown in Table \ref{statistics}. We opted using this data set because (1) it is publicly available; and (2) it allows a direct comparison with existing methods. 

We also use the data for distant supervision\footnote{We use the noisy but larger distantly supervised data to first adapt all models on the complaint classification task. Then we fine-tune them using the smaller original complaint data set.} collected by Preotiuc-Pietro et al. \shortcite{preotiuc2019automatically}. This extra `noisy' data source contains 18,218 complaint tweets collected by querying Twitter API with certain complaint related hashtags (e.g. \#badbusiness, \#badcustomerservice, etc.) and the same amount of non-complaint tweets that were sampled randomly.

\renewcommand{\arraystretch}{1.0}
\begin{table*}[!t]
\small
\begin{center}
\begin{tabular}{|l|c|c|}
\hline \rowcolor{dGray} \bf Domain & \bf Complaints & \bf Non-complaints \\ \hline
Food & 95 & 35 \\
\rowcolor{lGray} Apparel & 141 & 117 \\
Retail & 124 & 75 \\
\rowcolor{lGray} Cars & 67 & 25 \\
Services & 207 & 130 \\
\rowcolor{lGray} Software & 189 & 103 \\
Transport & 139 & 109 \\
\rowcolor{lGray} Electronics & 174 & 112 \\
Other & 96 & 33 \\ \hline
\rowcolor{lGray} Total & 1232 & 739 \\ \hline
\end{tabular}
\end{center}
\caption{\label{statistics} Data set statistics across 9 domains.}
\end{table*}

\section{Transformer-based Models}

 Transformer architectures trained on language modeling have been recently adapted to downstream tasks demonstrating state-of-the-art performance~\cite{weller2019humor,gupta2019effective,Maronikolakis2020}. In this paper, we adapt and subsequently combine transformers with external linguistic information for complaint prediction.
 
\paragraph{BERT, ALBERT and RoBERTa}
 Bidirectional Encoder Representations from Transformers (BERT)~\cite{devlin2018bert} learns language representations by jointly conditioning on both left and right contexts using transformers. It is trained on masked language modeling where some of the tokens are randomly masked with the aim to predict them using only the context. 
 
 We further experiment with ALBERT~\cite{lan2019albert} and RoBERTa~\cite{liu2019roberta}. ALBERT uses two parameter-reduction methods to address memory limitations and long training time of BERT: (a) factorized embedding parameterization; (b) cross-layer parameter sharing. RoBERTa is an extension of BERT trained on more data with larger batch size using dynamic masking (i.e. changeable masked tokens of each sequence during training epochs). We adapt BERT, ALBERT and RoBERTa by adding a linear layer with a sigmoid activation and then fine-tune it on the complaint classification data.
 
\paragraph{XLNet}
XLNet~\cite{yang2019xlnet} uses a similar architecture to BERT to learn bidirectional contextual information. Instead of masked tokens used in BERT, XLNet maximizes the expected log-likelihood of all possible factorization orders. We adapt and fine-tune the XLNet model for complaint prediction similar to BERT.
 
\paragraph{M-BERT}
  To combine our model with external linguistic information, we adapt the Multimodal BERT (M-BERT)~\cite{rahman2019m} model structure that has been introduced for multimodal modeling (text, image, speech). Instead of cross-modal interactions, we inject extra linguistic information as alternative views of the data into the pre-trained BERT model. We use (a) \textbf{Emotion}, a 9 dimensional vector obtained by quantifying six basic emotions of Ekman \shortcite{ekman1992argument} for each tweet using a predictive model by Volkova and Bachrach \shortcite{volkova2016inferring}; (b) \textbf{Topics}, a 200 dimensional vector representing word frequencies in word clusters designed to identify semantic themes in tweets by Preotiuc-Pietro et al. \shortcite{preoctiuc2015analysis,income}. To inject external linguistic information to M-BERT,\footnote{We experiment by injecting emotion ({\bf M-BERT - Emotion}) or topical information ({\bf M-BERT - Topics}) and their combination ({\bf M-BERT - Emotion+Topics})} we first project the linguistic information into vectors with similar size to the BERT CLS embeddings. Then we concatenate word representations obtained from BERT and the linguistic information (Emotion, Topics or Emotion+Topics) to generate combined embeddings. During concatenation, an Attention Gating Mechanism called Multimodal Shifting Gate~\cite{wang2019words} is applied to control the importance of each representation. Finally, the combined embeddings are fed to BERT for fine-tuning. The rest of the architecture is the same as BERT.
  
 

 \section{Experiments}
 
 \paragraph{Baselines}
 We compare the transformer-based models with two previous approaches for complaint identification by Preotiuc-Pietro et al. \shortcite{preotiuc2019automatically} and a transfer learning method: (1) Logistic Regression with bag-of-words trained using the original and distantly supervised complaint data ({\bf LR-BOW + Dist. Supervision}); (2) A Long-Short Term Memory  (\textbf{LSTM}) network~\cite{hochreiter1997long} that takes as input a tweet, maps its words to embeddings and subsequently passes them through the LSTM to obtain a contextualized representation which is finally fed to the output layer; (3) Adapting the pre-trained Universal Language Model Fine-tuning (\textbf{ULMFiT}) model~\cite{howard2018universal} for complaint prediction. ULMFiT uses a AWD-LSTM~\cite{merity2017regularizing} encoder for language modeling.

\paragraph{Hyper-parameters}
We use BERT, ALBERT and RoBERTa Base uncased models; fine-tuning them with learning rate $l$ = 1e-5, $l \in$ \{1e-4, 1e-5, 2e-5, 1e-6\}. We use the Base cased pre-trained XLNet tuning the learning rate over the same range as for BERT models. For ULMFiT, we use AWD-LSTM trained on Wikitext-103. We simplify the default fine-tuning by only unfreezing the last 1 layer, the last 2 layers and all layers with learning rates $l_1 = \frac{1e-4}{2.6^4}$, $l_2 = \frac{1e-4}{2.6^3}$ and $l_3$ = 1e-3 respectively. For M-BERT, the size of feature embeddings (Emotion, Topics and Emotion+Topics) is $h$ = 200, $h \in$ \{200, 400, 768\} with dropout $d$ = 0.1, $d \in \{.1, .5\}$ using the same parameters as BERT. The maximum sequence length is set to 49 covering 95\% of tweets in the training set.

\paragraph{Evaluation}
   Following Preotiuc-Pietro et al. \shortcite{preotiuc2019automatically}, we use a nested 10-fold cross-validation approach to conduct our experiments for complaint prediction. In the outer 10 loops, 9 folds are used for training and one for testing; while in the inner loops, a 3-fold cross-validation method is applied where 2 folds are used for training and one for validation. During training, an early stopping method is applied based on the validation loss. We measure predictive performance using the mean Accuracy, Precision, Recall and macro F1 over 10 folds (we also report the standard deviations).
 

\renewcommand{\arraystretch}{1.0}
\begin{table*}[!t]
\small
\begin{center}
\begin{tabular}{|l|c|c|c|c|}
\hline \rowcolor{dGray} \bf Model & \bf Acc & \bf P & \bf R & \bf F1\\ \hline
\rowcolor{mGray} {\bf Preotiuc-Pietro et al. \shortcite{preotiuc2019automatically}} &&&& \\ 
LR-BOW + Dist. Supervision  & 81.2 & - & - & 79.0\\
LSTM  & 80.2 & - & - & 77.0\\ \hline 
\rowcolor{mGray}  {\bf Transfer Learning Baseline} &&&& \\ 
ULMFiT & 82.4 $\pm$ .04 & 81.1 $\pm$ .04 & 81.8 $\pm$ .04 & 81.2 $\pm$ .05\\ 
ULMFiT + Dist. Supervision & 83.3 $\pm$ .05 & 82.5 $\pm$ .05 & 81.8 $\pm$ .04 & 81.9 $\pm$ .05\\ \hline
\rowcolor{mGray}  {\bf Transformers} &&&& \\   
BERT & $\bf 88.0$ $\pm$ .03 & $\bf 87.1$ $\pm$ .03 & $\bf 87.3$ $\pm$ .03 & $\bf 87.0$ $\pm$ .03\\ 
ALBERT & 85.9 $\pm$ .03 & 84.8 $\pm$ .03 & 84.6 $\pm$ .03 & 84.6 $\pm$ .03\\
RoBERTa & 87.6 $\pm$ .03 & 86.6 $\pm$ .03 & 86.9 $\pm$ .03 & 86.6 $\pm$ .03\\ 
XLNet & 83.9 $\pm$ .04 & 83.2 $\pm$ .04 & 82.3 $\pm$ .03 & 82.4 $\pm$ .05\\
M-BERT - Emotion & 87.3 $\pm$ .03 & 86.5 $\pm$ .04 & 86.0 $\pm$ .03 & 86.1 $\pm$ .04\\
M-BERT - Topics & 87.5 $\pm$ .03 & 86.7 $\pm$ .04 & 86.5 $\pm$ .03 & 86.4 $\pm$ .03\\
M-BERT - Emotion+Topics & 87.1 $\pm$ .03 & 86.4 $\pm$ .03 & 85.6 $\pm$ .03 & 85.9 $\pm$ .03\\
\hline

\hline

BERT + Dist. Supervision & 87.8 $\pm$ .03 & 87.0 $\pm$ .04 & 86.7 $\pm$ .03 & 86.7 $\pm$ .04\\ 
ALBERT + Dist. Supervision & 83.9 $\pm$ .04 & 82.6 $\pm$ .04 & 82.7 $\pm$ .04 & 82.6 $\pm$ .04\\
RoBERTa + Dist. Supervision & 85.2 $\pm$ .04 & 84.4 $\pm$ .05 & 84.0 $\pm$ .04 & 84.0 $\pm$ .04\\
XLNet + Dist. Supervision & 82.1 $\pm$ .05 & 81.7 $\pm$ .05 & 79.9 $\pm$ .05 & 80.1 $\pm$ .05\\
M-BERT - Emotion + Dist. Supervision & 87.7 $\pm$ .04 & 86.9 $\pm$ .04 & 87.2 $\pm$ .03 & 86.8 $\pm$ .04\\
M-BERT - Topics + Dist. Supervision & 87.6 $\pm$ .05 & 87.0 $\pm$ .05 & 86.9 $\pm$ .04 & 86.7 $\pm$ .05\\
M-BERT - Emotion+Topics + Dist. Supervision & 87.8 $\pm$ .04 & 87.1 $\pm$ .05 & 87.0 $\pm$ .04 & 86.9 $\pm$ .04\\\hline

\end{tabular}
\end{center}
\caption{\label{table1} Accuracy (Acc), Precision (P), Recall (R) and F1-Score (F1) for complaint prediction ($\pm$ std. dev.) Best results are in bold.}
\end{table*}


\paragraph{Predictive Results}

Table \ref{table1} shows results of the Transformer-based models as well as the baselines on the complaint prediction task.
All transformer-based models (BERT, ALBERT, RoBERTa and XLNet) perform better than the previous feature-based (LR-BOW + Dist. Supervision) and the non-transformer transfer learning baseline (ULMFiT), indicating a better capability on capturing idiosyncrasies of complaints syntax and semanatics. BERT outperforms other models overall across all metrics reaching a macro F1 up to 87, which is 8\% higher than the previous state-of-the-art~\cite{preotiuc2019automatically}. The results of RoBERTa are close to BERT with 86.6 macro F1 while ALBERT and XLNet achieve lower performance.  

Distant supervision is beneficial only to ULMFiT and M-BERT while BERT and other transformer models perform worse, which are consistent with results of Bataa and Wu \shortcite{bataa2019investigation} for sentiment analysis. Also, results of M-BERT models are comparable to BERT, among which M-BERT - Topics is slightly better with 86.4 macro F1. We notice that injecting external linguistic information in BERT's structure for fine-tuning does not help in our case without substantially hurting performance. We speculate that modifying BERT embeddings by injecting extra linguistic information is not complementary to BERT's text representations.


\paragraph{Error Analysis}
We also investigate the limitations in predicting capacity of our best performing model (BERT). We randomly analyze 100 cases in predictive results, where 50 cases were misclassified as non-complaints and another 50 cases were misclassified as complaints. In cases where complaints were misclassified as non-complaints, 26\% errors are due to implicit expressions while 14\% errors are because complaints contain irony. In the former situation, complaints express weak emotional intensity without explicit reproach, where complainers imply their dissatisfaction instead of directly complaining or mentioning the cause~\cite{trosborg2011interlanguage}. The following tweet is a typical example:
\begin{quote}
    \small
  \emph{It started yesterday , but I try again it could work normal. But since last night its just like this $<$url$>$}  
\end{quote}
Such expressions rarely include words related to complaints (e.g. `disappointed', `bad service') and are therefore difficult to be correctly classified. In the latter situation, complaints are expressed in an ironic way using terms such as `congratulations', `thank you' and `brilliant'. For instance, the following text was wrongly classified as a non-complaint:
\begin{quote}
    \small
  \emph{Thank you so much for making a box that shreds apart even when carried by both handles.}
\end{quote}
In cases where non-complaints were misclassified as complaints, errors can be roughly divided into four categories: (1) 26\% errors are because certain terms appear frequently in complaints during training such as `thank you', `dm', `lost', `work'. The following non-complaint was wrongly classified as a complaint:
\begin{quote}
    \small
  \emph{BTW $<$user$>$ -- $<$user$>$ did me right, and replaced my two failed batteries under warranty. I'm {\bf happy} :) {\bf thanks} $<$user$>$!}
\end{quote}
It contains similar words with the following complaint in the same fold (similarities highlighted in bold):  
\begin{quote}
    \small
  \emph{Was {\bf happy} to find out $<$user$>$ had an app to watch all their shows, until 6 episodes in it stops working. {\bf Thanks}! $<$user$>$}
\end{quote}
(2) 22\% errors due to interrogative tone, which is common in complaints. An example is \emph{``Folks , what is cost of text message to a us number?''} (3) 22\% errors are from negation words such as \emph{``No luck with pc or phone.''} (4) 12\% errors are because texts contain negative sentiment such as \emph{``This would be a terrible idea $<$url$>$''} are likely to be classified as complaints incorrectly since words such as `terrible' are widely used to express dissatisfaction. However, there are not enough cues to indicate violation of expectations. According to the statistics, the proportion of complaints misclassified as non-complaints (15.22\%) is higher than that of non-complaints misclassified as complaints (10.25\%) indicating implicit and figurative expressions as well as unknown factors in complaints are more challenging to identify.

\renewcommand{\arraystretch}{1.1}
\begin{table*}[!t]

\scriptsize
\begin{center}
\resizebox{\textwidth}{!}{
\begin{tabular}{|p{1cm}|cc|cc|cc|cc|cc|cc|cc|cc|cc|}
\hline \rowcolor{dGray}\bf Test
Train & \multicolumn{2}{c|}{F} & \multicolumn{2}{c|}{A} & \multicolumn{2}{c|}{R} & \multicolumn{2}{c|}{C} & \multicolumn{2}{c|}{Se} & \multicolumn{2}{c|}{So} & \multicolumn{2}{c|}{T} & \multicolumn{2}{c|}{E} & \multicolumn{2}{c|}{O}\\ \hline
Food & \multicolumn{2}{c|}{-} & \bf69.7 & 49.8 & \bf76.5 & 53.2 & \bf85.7 & 61.8 & \bf73.3 & 56.8 & \bf76.2 & 59.2 & \bf71.2 & 52.7 & \bf74.4 & 61.1 & \bf83.0 & 48.7\\ 
\rowcolor{lGray} Apparel & \bf75.0 & 69.7 & \multicolumn{2}{c|}{-} & 74.2 & \bf81.5 & \bf78.8 & 74.3 & 72.7 & \bf76.3 & 73.2 & \bf81.8 & 66.7 & \bf74.1 & 75.7 & \bf75.9 & 81.6 & \bf84.2\\ 
Retail & 74.9 & \bf75.9 & 72.8 & \bf80.0 & \multicolumn{2}{c|}{-} & \bf80.0 & 73.0 & 75.4 & \bf75.9 & 75.9 & \bf80.7 & 70.0 & \bf 74.9 & 72.7 & \bf75.3 & \bf86.6 & 79.5\\ 
\rowcolor{lGray} Cars & \bf76.1 & 57.1 & \bf70.2 & 62.1 & \bf75.5 & 65.1 & \multicolumn{2}{c|}{-} & \bf70.2 & 51.6 & \bf75.1 & 70.6 & \bf71.3 & 62.3 & \bf74.4 & 61.4 & \bf82.1 & 71.8\\ 
Services & \bf79.8 & 64.7 & 71.1 & \bf82.4 & \bf77.2 & 76.5 & \bf77.5 & 75.4 & \multicolumn{2}{c|}{-} & 73.8 & \bf78.6 & 73.4 & \bf76.3 & 75.5 & \bf79.4 & 78.9 & \bf83.0\\ 
\rowcolor{lGray} Software & \bf74.6 & 69.3 & 70.4 & \bf80.0 & 73.5 & \bf77.5 & \bf79.1 & 78.0 & \bf77.9 & 76.4 & \multicolumn{2}{c|}{-} & 73.4 & \bf75.2 & \bf76.7 & 76.2 & 81.7 & \bf82.0\\
Transport & \bf72.5 & 62.2 & 70.5 & \bf73.5 & 77.1 & \bf80.0 & \bf80.0 & 79.2 & 74.4 & \bf76.3 & \bf75.8 & 75.3 & \multicolumn{2}{c|}{-} & \bf72.4 & 70.4 & 82.0 & \bf82.6\\ 
\rowcolor{lGray} Electronics & 69.9 & \bf72.9 & 72.2 & \bf78.5 & 69.1 & \bf78.9 & 73.0 & \bf75.4 & 77.0 & \bf78.0 & 71.0 & \bf72.4 & \bf69.8 & 69.7 & \multicolumn{2}{c|}{-} & \bf82.1 & 80.8\\ 
Other & \bf65.9 & 64.8 & \bf75.2 & 74.8 & \bf79.2 & 72.2 & \bf81.7 & 69.2 & \bf76.3 & 69.9 & 76.5 & \bf77.9 & 70.6 & 70.6 & \bf72.8 & 70.8 & \multicolumn{2}{c|}{-}\\ \hline
All & 48.9 & \bf77.5 & 67.5 & \bf87.7 & 72.6 & \bf85.8 & 73.9 & \bf80.9 & 72.0 & \bf81.1 & 65.8 & \bf85.1 & 64.9 & \bf81.4 & 67.6 & \bf82.0 & 81.9 & \bf88.2\\ \hline 

\end{tabular} }

\end{center}

\caption{\label{table2}F1-score of models in Preotiuc-Pietro et al. \shortcite{preotiuc2019automatically} (left) and BERT (right) trained from one domain and tested on other domains. Domains include Food (F), Apparel (A), Retail (R), Cars (C), Services (Se), Software (So), Transport (T), Electronics (E) and Other (O). The All line shows results on training on all categories except the category in testing. Best results are in bold.}
\end{table*}

\paragraph{Cross Domain Experiments}

Finally, we use BERT to train models on one domain and test on another as well as training on all domains except the one that the model is tested on. Table \ref{table2} shows the performance of models in Preotiuc-Pietro et al. \shortcite{preotiuc2019automatically} (left) and BERT (right) across 9 domains. We first observe that BERT results in nearly half of the cases when training on a single domain are lower than LR-BOW (especially `Food', `Car' and `Other') while BERT trained on all domains performs better across all testing domains, achieving a macro F1 up to 88.2 when tested on `Other'. This indicates that, fine-tuning BERT on a small training data set (`Food', `Car' and `Other' are three of the domains with the smaller amount of data) is not enough to make it perform well. In contrast, it achieves better performance consistently on larger data sets (All). We also notice that BERT performs robustly for domain pairs where the domains are either used for training or testing. For example, training on `Apparel' achieves high performance when testing on `Software' (81.8 F1) and vice versa (80.0 F1). Furthermore, domain relevance affect predictive performance. For example, BERT trained on `Transport' achieves 79.2 F1 when tested on `Car', which is the highest performance compared to other training domains since these two domain share common vocabulary (see `Car' column for BERT).

\section{Conclusion}
We evaluated a battery of transformer networks on the Twitter complaint identification task and obtained 87 macro F1, which outperforms the previous state-of-the-art results of Preotiuc-Pietro et al. \shortcite{preotiuc2019automatically}. We further presented a thorough analysis of the limitations of our models in predicting complaints. In future work, we intend to explore more in how we can combine other sources of linguistic information with transformers as well as information from other modalities such as images.

\section*{Acknowledgements}
Nikolaos Aletras is supported by ESRC grant ES/T012714/1.

\bibliographystyle{coling}
\bibliography{coling2020}

\end{document}